\documentclass[runningheads]{llncs}
\usepackage{graphicx}
\usepackage{graphicx}
\usepackage{diagbox}
\usepackage[breaklinks,colorlinks,bookmarks=false]{hyperref}
\usepackage{amsmath}
\usepackage{amssymb}
\usepackage[labelformat=simple]{subcaption}
\usepackage{multirow}
\usepackage{multicol}
\usepackage{diagbox}
\usepackage{cite}

\makeatletter
\usepackage{xspace}
\DeclareRobustCommand\onedot{\futurelet\@let@token\@onedot}
\def\@onedot{\ifx\@let@token.\else.\null\fi\xspace}
\def\etal{\emph{et al}\onedot}
\makeatother

\begin{document}
\title{Multi-model learning by sequential reading of untrimmed videos for action recognition}
%
%
\author{
Kodai Kamiya\inst{1} \and
Toru Tamaki\inst{1}\orcidID{0000-0001-9712-7777}}
\authorrunning{K. Kamiya et al.}
%
\institute{
Nagoya Institute of Technology, Nagoya, Japan\\
\email{k.kamiya.865@nitech.jp},
\email{tamaki.toru@nitech.ac.jp}
}
\maketitle              

\begin{abstract}
We propose a new method for learning videos by aggregating multiple models by sequentially extracting video clips from untrimmed video. The proposed method reduces the correlation between clips by feeding clips to multiple models in turn and synchronizes these models through federated learning. Experimental results show that the proposed method improves the performance compared to the no synchronization.

\end{abstract}

\section{Introduction}

In recent years, there has been a lot of research on video recognition
for various potential applications in the real world such as 
action recognition
\cite{Hara_IEICE_ED2020_Action_Recognition_Survey,Kong_IJCV2022_Action_Recognition_Survey,Ulhaq_arXiv2022_Transformers_Action_Recognition_Survey,Zhu_arXiv2020_Action_Recognition_Survey}, 
action segmentation,
temporal \cite{Xia_IEEEAccess2020_Temporal_Action_Localization_survey} and spatio-temporal action localization \cite{Bhoi_arXiv2019_Spatio-Temporal_Action_Recognition_Survey}.
For these tasks, there are two types of videos: \emph{trimmed} and \emph{untrimmed} videos.
\emph{Trimmed videos} are relatively short videos, trimmed from the original videos, usually in a few or several seconds \cite{kay_arXiv2017_kinetics400,Soomro_arXiv2012_UCF101}. Typically, an action label is assigned to each video to perform action recognition tasks. 
On the other hand, \emph{untrimmed video} refers to videos that have not been trimmed,%
\footnote{
In fact, untrimmed videos were 
actually trimmed by those who uploaded the videos,
but not by those who annotate them.}
and the length of a single video can range from several to tens of minutes \cite{Gu_2018CVPR_AVA-Actions,Rohrbach_CVPR2012_MPII_Cooking}. Therefore, the content of an untrimmed video is complex and is usually used for action segmentation or (spatio)temporal localization tasks that require frame-level annotation rather than video-level annotation. 

Although many methods have been proposed for the action recognition task of classifying trimmed videos \cite{Tran_2015ICCV_C3D, Carreira_2017CVPR_I3D, Feichtenhofer_2020CVPR_X3D,Feichtenhofer_2019ICCV_SlowFast,Zhang_ACMMM2021_TokenShift,Arnab_2021_ICCV_ViVit,Bertasius_ICML2021_TimeSformer}, they all share a common procedure for handling input videos: a specified number of frames $T$ at a fixed frame interval (stride) $s$ are extracted from a single video, and a stack of these frames (often called ``clips'') is input into the model; for example, 16 frames with stride of 5 frames \cite{Feichtenhofer_2020CVPR_X3D}, 8 frames with stride of 32 frames \cite{Zhang_ACMMM2021_TokenShift}, etc. In this way, the input to the model is a three-dimensional tensor so that these models can be handled in the same way.

\begin{figure}[t]
  \centering

  \begin{minipage}[b]{0.46\linewidth}
    \centering
    \includegraphics[width=\linewidth]{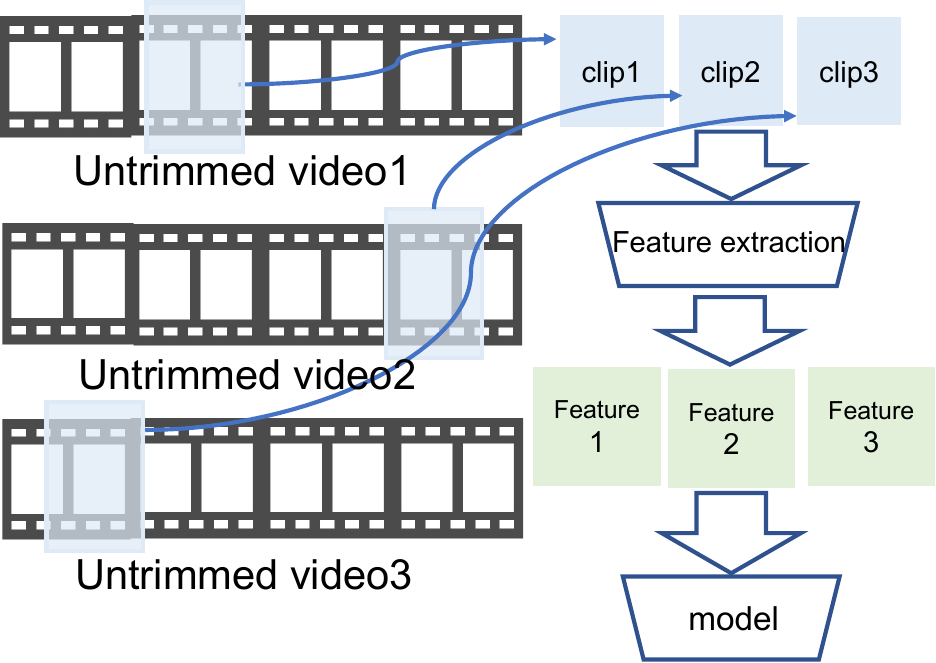}
    \subcaption{}
    \label{fig:sampling_prior_work}
  \end{minipage}
  \hfill
  \begin{minipage}[b]{0.48\linewidth}
    \centering
    \includegraphics[width=\linewidth]{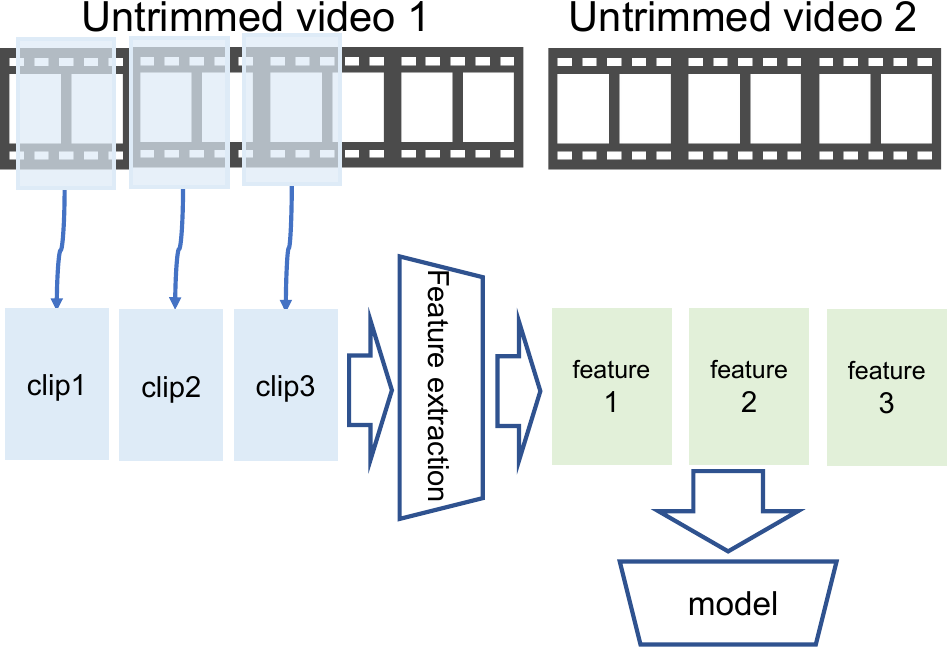}
    \subcaption{}
    \label{fig:sampling_proposed_method}
  \end{minipage}

  \caption{Two ways for loading video clips from untrimmed videos.
    (a) A common way is to split untrimmed videos into clips for pre-computed features.
    (b) Another way is reading untrimmed video from the beginning to extract clips sequentially.
  }
  \label{fig:overview}
\end{figure}

On the other hand, ``untrimmed video'' tasks need to deal with various actions in a single video, and the length of the video makes it difficult for the model to use the video directly. Therefore, the handling of untrimmed videos varies, but in most cases (Fig.~\ref{fig:sampling_prior_work}), untrimmed videos are divided into multiple trimmed videos (or clips) of fixed or variable length, and action features of each clip are then precomputed by using action recognition models \cite{Carreira_2017CVPR_I3D,Tran_2015ICCV_C3D} as a feature extractor \cite{Feichtenhofer_2019ICCV_SlowFast, Ghadiyaram_CVPR2019_LargeScale, Duan_ECCV2020_omni}.
This somehow avoids the drawback of the length issue of untrimmed videos.

However, the feature extractor is usually fixed and not end-to-end fine-tuned. In general, training of the model in an end-to-end manner, including the feature extractor, is expected to improve the performance of tasks involving untrimmed videos. Therefore, in this work, we consider directly handling untrimmed videos.

When training an action recognition model, usually trimmed videos in the training set are randomly selected and clips are extracted from random locations in the selected trimmed video, as shown in Fig.~\ref{fig:sampling_prior_work}, in order to ensure that sampling is expected to be i.i.d. However, this is not an efficient way for untrimmed videos because random access to video files in the storage occurs, as well as seeking to a random position in the video. This is not a problem for short trimmed videos,%
\footnote{Nevertheless, seeking to random frames in a video file is time-consuming, then a common practice is to extract all frames and stores as JPEG files in advance\cite{Otani_IEEEAccess2022_MPEG_JPEG}.
}
but it is very inefficient for long untrimmed videos.

A naive solution would be to read the untrimmed video sequentially and extract clips from the beginning on the fly during training \cite{Kondratyuk_2021CVPR_MoViNets,Wu_CVPR2022_MemViT} as shown in Fig.~\ref{fig:sampling_proposed_method}.
However, this sampling is non-iid and the clips are similar (highly correlated), which hinders the model to be efficiently trained. Therefore, it is necessary to reduce the correlation between the clips without losing efficiency.

To do this, we propose using
federated learning \cite{Zhu_Neurocomputing2021_Federated_Learning_survey}
to train replicated multiple models with clips that are sequentially extracted from untrimmed videos.
Federated learning is a type of machine learning in which multiple models are synchronized in a distributed training environment. 
The proposed method reduces the correlation between clips by feeding clips to multiple models in turn and synchronizes these models through federated learning.
This enables end-to-end learning by handling untrimmed video without precomputation of clip features.

\section{Related Work}

\subsection{Action Recognition}

For action recognition, the task of predicting a category at video level
\cite{Hara_IEICE_ED2020_Action_Recognition_Survey,Kong_IJCV2022_Action_Recognition_Survey,Ulhaq_arXiv2022_Transformers_Action_Recognition_Survey,Zhu_arXiv2020_Action_Recognition_Survey},
and many models have been proposed,
including CNN-based \cite{Carreira_CVPR2017_I3D,Feichtenhofer_2020CVPR_X3D,Feichtenhofer_2019ICCV_SlowFast}
and Transformer-based
\cite{Arnab_2021_ICCV_ViVit, Bertasius_ICML2021_TimeSformer, Zhang_ACMMM2021_TokenShift},
as well as
many available datasets
\cite{kay_arXiv2017_kinetics400,Soomro_arXiv2012_UCF101,Kuehne_ICCV2011_HMDB51,Goyal_2017ICCV_ssv2}.
These methods all share the same procedure; taking a video clip consisting of frames extracted from a single video as input.
Typically, a few seconds of the trimmed video
is extracted as a clip consisting of several frames with a stride between frames,
even when the untrimmed videos are in several seconds
\cite{kay_arXiv2017_kinetics400,Goyal_2017ICCV_ssv2}.

In our work, we use two trimmed and untrimmed
action recognition datasets;
UCF101 \cite{Soomro_arXiv2012_UCF101},
HMDB51 \cite{Kuehne_ICCV2011_HMDB51}, and
MPII Cooking\cite{Rohrbach_CVPR2012_MPII_Cooking}.
UCF101 has trimmed videos in 7.21 seconds on average, but includes videos more than one minute; hence, it would be suitable for our work.
MPII Cooking is a dataset of untrimmed videos of several minutes, and has multiple labels assigned to a single video.

\subsection{Effects of data shuffling}

It is common to randomly sample data from a training set (usually called shuffling) \cite{DeepwizAI_blog2021}, and even when training samples are loaded sequentially, a pseudo-shuffling with a queue (or shuffle buffer) is used
\cite{Aizman_arXiv2020_WebDataset,TorchData_github2022,WebDataset_github2020}.
However, the impact of shuffle on performance has not been studied systematically well. Nguyen \etal \cite{Nguyen_IPDPS2022_Globally_Re-shuffle} compared global shuffling (all data are exchanged with all clients) and partial shuffling (only a portion of the data is exchanged) when local data is exchanged with other clients in a distributed environment. They reported that partial shuffling does not significantly affect performance.

On the other hand, in this study, we assume a situation where multiple models are trained in parallel on highly correlated data, i.e., clips continuously extracted from untrimmed video, so shuffling between clients (between models) is not possible. In experiments, we compare the performance impact and efficiency of the proposed method with the case where video clips are randomly loaded.
Recently proposed MemViT \cite{Wu_CVPR2022_MemViT} and MoViNets \cite{Kondratyuk_2021CVPR_MoViNets}
aim at online learning and inference of video clips continuously extracted from untrimmed video,
but do not address the issue of correlation between successive video clips.

\subsection{Learning with long video}

Efforts to learn long, untrimmed videos have been studied from various perspectives.
Yang \etal
\cite{Yang_CVPR2021_beyond_short_clips}
proposed collaborative memory, which integrates clip-level learning with video-level learning.
Pang \etal \cite{Pang_CVPR2021_pgt}
proposed progressive training, which introduces a Markov model to learn temporally adjacent clips.
However, these predict the video-level category for trimmed videos in the action recognition task, not for untrimmed videos.
They used I3D \cite{Carreira_CVPR2017_I3D},
ResNet-3D \cite{Hara_CVPR2018_ResNet_3D}
and SlowFast \cite{Feichtenhofer_2019ICCV_SlowFast}
to extract clip features
and trained end-to-end including these feature extractors.
However, they do not learn directly from untrimmed videos, but read trimmed videos that have been cut out in advance before training.

Qing \etal
\cite{Qing_CVPR2022_learning_from_untrimmed_video}
proposed HiCo, which uses the visual and topic consistency of untrimmed videos. In their implementation, however, untrimmed videos are cut and saved as clips in an offline process. Although this can speed up the seek to random positions in untrimmed videos, it is completely different from learning by continuous clip extraction, which is our goal.

\section{Method}

In this section, we describe the process of creating clips from untrimmed video and training multiple models in parallel. 

\subsection{Data}

Let $V = \{v_0, v_1, \ldots, v_{N-1}\}$ be a video dataset, where each $v_i \in \mathbb{R}^{T_i \times 3 \times H \times W}$ is an untrimmed video with $T_i$ frames. $H, W$ are the height and width of the frame, assumed to be the same for all videos.%
\footnote{%
It is assumed that video files are either transcoded in preprocessing or resized when frames are acquired from video files. Also, the frame rate is assumed to be the same for all videos.
}

Let $\ell_i \in \{1, \ldots, C\}^{T_i}$ be the frame-by-frame annotation for video $v_i$ where $C$ is the number of categories. That is, $\ell_i(t)$ is the label for video $v_i(t)$, which is frame $t$ of video $i$.

\subsection{Creating clips}

Let $x_{i,n} \in \mathbb{R}^{T \times 3 \times H \times W}$ be a clip continuously extracted from video $v_i \in \mathbb{R}^{T_i \times 3 \times H \times W}$ by
\begin{align}
    x_{i,n}(t) = v_i(t - s_{i,n}), \quad t = 0, \ldots, T-1,
\end{align}
where $s_{i,n}$ is the start frame of $n$-th clip ($n = 0, 1, \ldots , N_i - 1$) and $s_{i,0} = 0$.
$T$ is the number of frames in the clip, which is the same for all clips. Since the total number of frames $T_i$ differs from video to video, $N_i = \lfloor T_i / T \rfloor$ is also different for each video.

To simplify the problem, we assume that the category labels of the frames in a clip are the same and identical to $\ell_{i,n}$;
\begin{align}
\ell_{i,n} = \ell_i(t), \quad \forall t = s_{i,n}, s_{i,n} + 1, \ldots, s_{i,n} + T-1.
\end{align}
Clips that do not satisfy this condition are not used simply by discarding the clip, and the procedure continues by moving the start frame of the clip $s_{i,n}$ to the frame when the label is changed. Therefore, the total number of clips extracted from a video $v_i$ is $N_i \le \lfloor T_i / T \rfloor$.%
\footnote{%
Note that the above procedure assumes that consecutive frames are used to create a clip (i.e., the unit stride)
and a clip has a single label,
however, it is easy to extend our work to non-unit stride clip generations and frame-level annotations.}

The resulting training data is a sequence of pairs
of video clips and their labels;
$$
(x_{i,n}, \ell_{i,n})_{i=0,\ldots,N-1, n=0, \ldots, N_i-1}.
$$
When clips are extracted sequentially from multiple videos, the sequence is as follows;
$$
(x_{0,0}, \ell_{0,0}),
(x_{0,1}, \ell_{0,1}),
\ldots,
(x_{0,N_0-1}, \ell_{0,N_0-1}), 
(x_{1,0}, \ell_{1,0}),
(x_{1,1}, \ell_{1,1}),
\ldots
$$
and so on.
For convenience, we use $j=0,1,\ldots,J$
to denote the index of this order,
and 
the training sample sequence is then simply denoted by
$$
(x_0, \ell_0),
(x_1, \ell_1),
\ldots,
(x_j, \ell_j),
\ldots
$$
and so on.

\subsection{Model input}

The successive clips in the training data sequence created as above
are highly correlated with each other.
The idea of the proposed method is to use multiple models and alternate the use of consecutive clips to decrease the correlation of training clips for a single model.

\subsubsection{Training in a normal case.}

Let $f$ be a model to be trained and $w$ be its parameters. Suppose that $f$ takes an input $x_j$ and outputs a category prediction $\hat{y}_j \in [0,1]^C$. To minimize the loss of cross-entropy
\begin{align}
    L_\mathrm{CE} 
    = E\left[ L_\mathrm{CE}( \hat{y}_{j}, \ell_j )  \right]
    = \frac{1}{J} \sum_{j=0}^J L_\mathrm{CE}( \hat{y}_{j}, \ell_j ),
\end{align}
we compute 
the average of the loss for each sample $x_b$ in a batch of batch size $B$,
\begin{align}
    L_{\mathrm{CE}_{b}} = L_\mathrm{CE}(\hat{y}_{b}, \ell_b), \quad
    L_\mathrm{CE} = \frac{1}{B} \sum_{b=0}^{B-1} L_{\mathrm{CE}_{b}},
\end{align}
where $\hat{y}_{b}, \ell_b$ is the prediction for $x_b$ and the true value.
Then we update the parameter
with a learning rate of $\eta$;
\begin{align}
    w \leftarrow w + \eta \nabla L_\mathrm{CE}.
\end{align}

\begin{figure}[t]
  \centering

  \includegraphics[width=.6\linewidth]{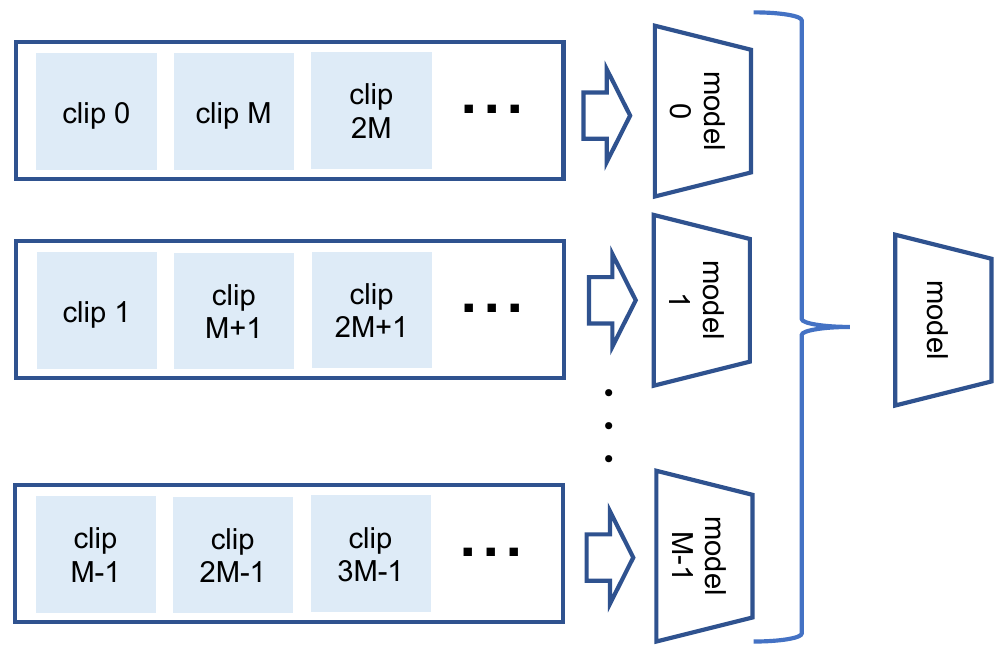}

  \caption{The proposed method for reading video clips with multiple models.}
  \label{fig:multiple_models_dataload}
\end{figure}

\subsubsection{Training in our case.}

In our work, $M$ replicas of $f$ with different initial values are prepared, with $f_m$ as the $m$-th model and $w_m$ as its parameters ($m=0,\ldots,M-1$).
Then, the samples in the batch are input to each model in turn as follows (see Fig.~\ref{fig:multiple_models_dataload});
\begin{align}
    \hat{y}_{b,0} &= f_0(x_b), \quad b=0, M, 2M, \ldots \\
    \hat{y}_{b,1} &= f_1(x_b), \quad b=1, 1+M, 1+2M, \ldots \\
    \vdots \nonumber\\
    \hat{y}_{b,m} &= f_m(x_b), \quad b=m, m+M, m+2M, \ldots \\
    \vdots \nonumber\\
    \hat{y}_{b,M-1} &= f_{M-1}(x_b), \quad b=M-1, 2M-1, 3M-1, \ldots
\end{align}
For simplicity,
the batch size $B$ is assumed to be a multiple of
the number of models $M$.
If this is not the case, the order of the model
is randomly shuffled for every batch to ensure that all models use
the same number of samples on average.

The above process is equivalent to defining the loss for each model $f_m$ by
\begin{align}
    L_{\mathrm{CE}_{b,m}} &= L_\mathrm{CE}(\hat{y}_{b,m}, \ell_b) \\
    L_{\mathrm{CE}_{m}} &= \frac{1}{B/M} \sum_{\substack{b=m,\\ m+M,\\ m+2M, \ldots}}^{B-M+m} L_{\mathrm{CE}_{b,m}}
\end{align}
and updating each model as follows;
\begin{align}
    w_m & \leftarrow w_m + \eta \nabla L_{\mathrm{CE}_{m}}.
\end{align}

In this way, the training sample sequence for each model is
$$
(x_m, \ell_m),
(x_{m+M}, \ell_{m+M}),
(x_{m+2M}, \ell_{m+2M}),
\ldots
$$
and so on.
If $M$ is large so that the autocorrelation of the training sample sequence is low enough, it is expected that
the problem of high correlation will be mitigated.

\subsection{Model synchronization by federated learning}

Multiple models trained in parallel need to be aggregated. In the following, we propose the use of federated learning.
A simple one is
FedAvg \cite{McMahan_AISTATS2017_federated_averaging},
which averages parameters of models after each model update for a given batch and reassigned the parameters to each model;
\begin{align}
    w_m & \leftarrow w_m + \eta \nabla L_{\mathrm{CE}_{m}}, \quad m = 0, \ldots, M-1 \\
    \bar{w} & \leftarrow \frac{1}{M} \sum_{m=0}^{M-1} w_m \\
    w_m & \leftarrow \bar{w}, \quad m = 0, \ldots, M-1.
\end{align}

FedAvg synchronizes the parameters of all models at each update, which leads to all models being similarly biased. However, this is essentially the same as using a single model.
Therefore, we instead partially synchronize parameters by moving average as in FedProx \cite{Tian_MLSYS2020_FedProx};
\begin{align}
    w_m & \leftarrow (1 - \alpha_m) \bar{w} + \alpha_m w_m, \quad m = 0, \ldots, M-1,
    \label{eq:fedprox update}
\end{align}
where $\alpha_m \in [0, 1]$ is momentum of each model.
It can be fixed during training; however,
by scheduling $\alpha_m$ closer to 0 as training progresses, the parameters of all models are gradually synchronized at the end of training.

When multiple models trained as above are used for validation, we merge them to generate a single model.

\section{Experiments}

To evaluate the proposed method, experiments were conducted on several datasets to evaluate the performance of the proposed method and to compare it with the conventional method.

\subsection{Experimental Setup}

\subsubsection{Trimmed video datasets}

UCF101 \cite{Soomro_arXiv2012_UCF101} is a dataset for action recognition of 101 classes of human actions, consisting of a training set of 95k videos and a validation set of 35k videos.
An action category is annotated per video,
hence each video is considered as trimmed,
while the video length is from 1.06 to 71.04 seconds with the average of 7.21 seconds.

HMDB51 \cite{Kuehne_ICCV2011_HMDB51} 
is a dataset for action recognition,
consisting of 3.6k videos in the training set and 1.5k videos in the validation set, with video-level annotation of 51 human action categories.
The shortest video is 0.63 seconds and the longest is 35.43 seconds, with an average length of 3.15 seconds.

\subsubsection{Trimmed video dataset}

MPII Cooking \cite{Rohrbach_CVPR2012_MPII_Cooking} is a dataset that includes 44 untrimmed videos of 12 subjects (s08 -- s20) who cook 14 different dishes in a kitchen. Videos vary in length from 3 to 40 minutes, with a total of 8 hours of footage. For each frame, an action label was assigned to one of 65 cooking activities. Each of the 5609 action intervals is assigned a single label, and the official split divides them into training and validation sets. However, there is an issue with the official split for our experiment as the training and validation intervals are both exist in the same untrimmed video.
Thus, we created our own train-validation split as follows;
\begin{itemize}
    \item train set: videos of the first 8 subjects (s08 -- s16)
    \item validation set: videos of the remaining 4 subjects (s17 -- s20)
\end{itemize}
This was used for sequential sampling (see below).
For random sampling (see also below), we cut and saved each annotated action interval as a trimmed video with an action label.
This results in 3774 trimmed videos for training and
1835 trimmed videos for validation.

\subsubsection{Clip extraction}

In experiments, we compare two types of clip extraction.

\begin{itemize}

\item \textbf{random clip sampling} (Fig.~\ref{fig:sampling_prior_work}):
First, we randomly select one of the trimmed videos from the training set.
From the selected video, we randomly specify a start frame and extract consecutive frames corresponding to the specified clip duration in seconds.
From there, $T=16$ frames are sampled uniformly in the time direction to create a clip.
In experiments, the duration of the clip was set to 64 / 15 = 2.56 seconds.

\item \textbf{sequential clip sampling} (Fig.~\ref{fig:sampling_proposed_method}):
First, an untrimmed video is randomly selected from the training set.
From the first frame, a specified number of consecutive frames are extracted at a specified stride $s$ (the number of frames to the next frame) to create a single clip.
In this case, $T=16$ frames with stride $s=1$ are used as a clip.    
\end{itemize}

Videos from UCF101 and HMDB51 are considered trimmed for random clip sampling, but are regarded untrimmed for sequential sampling.
For MPII, we used the trimmed video sets for random clip sampling, while the original untrimmed videos were used for sequential sampling.

\subsubsection{Training and inference}

For training, the short sides of the frames were randomly determined within a range of $[256, 320]$ pixels and resized while preserving the aspect ratio. Then, $224 \times 224$ pixels at random locations were cropped and flipped horizontally with a probability of $1/2$.
For validation, clips were generated in the same way for both the sequential and random cases. For the random case, each frame was resized to 256 pixels on the shorter side, while preserving the aspect ratio.

The optimizer was SGD with a learning rate of 1e-3, weight decay 5e-5, momentum 0.9.
The same settings were used for multiple models.
The batch size was set to $B=8$ and the models were trained for 5 epochs.

All synchronization momentum weights of the $M$ models were set to be identical (that is, $\alpha_1 = \cdots = \alpha_M$) during training.
For inference, the parameters of the $M$ models were averaged to generate a single model, which was used for validation.

\subsubsection{Models}

We used X3D-M \cite{Feichtenhofer_2020CVPR_X3D}, a lightweight 3D CNN-based action recognition model,
pre-trained on Kinetics400 \cite{kay_arXiv2017_kinetics400}.
When $M \ge 2$, we prepared multiple X3D-M instances
with differently initialized heads.

\subsection{Results}


\subsubsection{Synchronizing model parameters with $\alpha$.}

First, Table \ref{tab:alpha_vs_performance} shows the performance of UCF101 and HMDB51 for different values of synchronization momentum $\alpha_m$ in update Eq.\eqref{eq:fedprox update}
when two models were used ($M=2$).
$\alpha = 0$ means that the parameters of all models are synchronized after each iteration of the training, while $\alpha = 1$ means that each model is trained separately without parameter synchronization.
The performance was better when $\alpha = 0.2 \sim 0.4$,
therefore we will use $\alpha = 0.3$ in the following experiments.

We also linearly increased $\alpha$ by 0.2 per epoch from 0.0 to 1.0, or decreased by 0.2 per epoch from 1.0 to 0.0. 
Higher performance was achieved when $\alpha$ increased from 0,
i.e., gradually learning the parameters of the models separately ($\alpha$ to 1.0) from completely synchronized ($\alpha = 0.0$).
This is not intuitive; hence we will investigate these cases with more than two models.

\begin{table}[t]
    \centering
    \caption{The top-1 performance with diﬀerent values of $\alpha$ when $M=2$.
    For last two columns $\alpha$ was increased or decreased by 0.2 at every epoch.}
    \label{tab:alpha_vs_performance}
    \begin{tabular}{c|cccccc|cc}
    $\alpha$    & 0.0   & 0.2   & 0.4   & 0.6   & 0.8   & 1.0   & 0.0$\sim$1.0 & 1.0$\sim$0.0 \\ \hline
    UCF  & 96.59 & 96.33 & 96.72 & 95.90 & 96.72 & 96.53 & 97.04 & 96.72 \\
    HMDB & 74.14 & 75.42 & 76.41 & 75.82 & 76.54 & 75.95 & 76.80 & 73.14 
\end{tabular}
\end{table}

\subsubsection{Using $M$ models.}

Next, we compare the performance for different $M$, the number of models. Table \ref{tab:performance_with_M_models} shows that in the case of sequential sampling, the performance increases until $M=3$, but having more than three models had a negative effect on the performance as when $M=4$. This trend is not observed in the case of random sampling, where the performance decreases as the number of models increases. 
Sequential sampling shows the effectiveness,
although more experiments are needed with a larger number of models.


\begin{table}[t]
    \centering

    \caption{Performance comparison of synchronized multiple models.
    $M=1$ stands for no synchronization.}
    \label{tab:performance_with_M_models}
    
    \begin{tabular}{c|ccc|ccc}
        & \multicolumn{3}{c|}{random}                                    & \multicolumn{3}{c}{sequential}                               \\
    $M$   & UCF   & HMDB  & MPII  & UCF   & HMDB  & MPII \\ \hline
    1   & 95.29 & 75.33 & 45.12 & 97.92 & 77.98 & 27.80 \\
    2   & 95.58 & 76.33 & 43.53 & 97.47 & 77.38 & 25.83 \\
    3   & 94.57 & 75.33 & 45.50 & 97.02 & 77.68 & 33.92 \\
    4   & 95.02 & 74.73 & 43.42 & 96.13 & 76.56 & 29.14 \\
    \end{tabular}
    \end{table}



\begin{table}[t]
    \centering

    \caption{Comparisons of computation time per iteration (in seconds) including data loading, forward and backward computation.}
    \label{tab:computation speed}

    \begin{tabular}{c|ccr|ccc}
        & \multicolumn{3}{c|}{random} & \multicolumn{3}{c}{sequential} \\
    $M$ & UCF  & HMDB  & MPII  & UCF  & HMDB & MPII \\ \hline
    1   & 1.22 & 1.26  & 13.43 & 0.90 & 1.31 & 8.84 \\
    2   & 1.25 & 1.31  & 13.74 & 0.96 & 1.30 & 9.22 \\
    3   & 1.29 & 1.35  & 13.84 & 1.01 & 1.59 & 8.87 \\
    4   & 1.35 & 1.40  & 13.94 & 1.16 & 2.79 & 9.49
    \end{tabular}
    \end{table}

\subsubsection{Comparison of efficiency.}

Table \ref{tab:computation speed} shows the average computation time for one iteration in each experimental setting. This includes data loading from trimmed or untrimmed videos for clip sampling and model forward and backward computation.
It can be seen that the computation time is shorter for sequential than for random. However, the speed of data loading may be affected by many factors, and we will investigate the efficiency more in depth in the future.

\section{Conclusion}

In this paper, we propose a method for learning multiple models with synchronization of federated learning and sequential clip sampling that sequentially extracts video clips from untrimmed video. Experimental results show that a single model merged from multiple trained models with synchronization improves performance compared to a model
without synchronization.
Future work includes optimizing the code to further improve the efficiency of data loading and synchronization when more models are involved.

\section*{Acknowledgements}
This work was supported in part by JSPS KAKENHI Grant Number JP22K12090.

\bibliographystyle{splncs04}
\bibliography{all,mybib}

\end{document}